\documentclass[10pt,twocolumn,letterpaper]{article}
\usepackage{iccv}
\usepackage{times}
\usepackage{epsfig}
\usepackage{graphicx}
\usepackage{amsmath}
\usepackage{amssymb}
\usepackage{color}
\usepackage{algorithm}
\usepackage{algorithmic}
\usepackage{multirow}
\usepackage[bold]{hhtensor}
\usepackage{subfig}
\newcommand{\squishlist}{
 \begin{list}{$\bullet$}
  { \setlength{\itemsep}{0pt}
     \setlength{\parsep}{1pt}
     \setlength{\topsep}{1pt}
     \setlength{\partopsep}{0pt}
     \setlength{\leftmargin}{1.5em}
     \setlength{\labelwidth}{1em}
     \setlength{\labelsep}{0.5em} } }

\newcommand{\squishend}{
  \end{list}  }
\usepackage[pagebackref=true,breaklinks=true,letterpaper=true,colorlinks,bookmarks=false]{hyperref}
\iccvfinalcopy 

\ificcvfinal\fi
\begin{document}

\title{Detecting 11K Classes: Large Scale Object Detection without Fine-Grained Bounding Boxes}

\author{Hao Yang \qquad Hao Wu \qquad Hao Chen \\Amazon Web Services \\\{haoyng, goodwu, hxen\}@amazon.com}
\maketitle

\begin{abstract}
Recent advances in deep learning greatly boost the performance of object detection. State-of-the-art methods such as Faster-RCNN, FPN and R-FCN have achieved high accuracy in challenging benchmark datasets. However, these methods require fully annotated object bounding boxes for training, which are incredibly hard to scale up due to the high annotation cost. Weakly-supervised methods, on the other hand, only require image-level labels for training, but the performance is far below their fully-supervised counterparts. In this paper, we propose a semi-supervised large scale fine-grained detection method, which only needs bounding box annotations of a smaller number of coarse-grained classes and image-level labels of large scale fine-grained classes, and can detect all classes at nearly fully-supervised accuracy. We achieve this by utilizing the correlations between coarse-grained and fine-grained classes with shared backbone, soft-attention based proposal re-ranking, and a dual-level memory module. Experiment results show that our methods can achieve close accuracy on object detection to state-of-the-art fully-supervised methods on two large scale datasets, ImageNet and OpenImages, with only a small fraction of fully annotated classes.
\end{abstract}


\vspace{-0.2in}
\section{Introduction}
Object detection has been in the center stage of computer vision research, and the recent development of deep learning has vastly improved its performance. State-of-the-art algorithms, such as R-FCN~\cite{Dai2016}, FPN~\cite{Lin2017} and Mask-RCNN~\cite{He2017} have achieved high accuracy on challenging benchmark datasets. However, these methods rely on accurate and complete annotations of object bounding boxes, which is expensive and time consuming to collect.

Exhaustively annotating bounding boxes for a large scale dataset with fine-grained level labels (here by large scale, we not only mean large number of instances $N$, but also large number of categories $C$) is extremely laborious. Due to the multi-label nature of detection problem, the annotation costs is nearly $O(NC)$. 
Even with state-of-the-art annotating methods~\cite{Papadopoulos2016, Papadopoulos2017}, it is still much more costly than annotating image level labels. Therefore, existing object detection datasets are either small scale and relative fully annotated, such as PASCAL VOC~\cite{Everingham2010} and MS COCO~\cite{Lin2014}, or large scale but only a part of the categories are annotated with bounding boxes, such as ImageNet~\cite{Deng2009} and OpenImages~\cite{openimages}, and even for those categories with bounding boxes, missing labels are common~\cite{Wu2018}.

Compared to annotating bounding boxes, annotating image level labels is cheaper. For example, ImageNet has $11$K categories that are consider trainable (more than $500$ training images) with $12$M total training images, but out of theses images, only $3$K categories and below $1$M training images have bounding boxes annotations~\cite{RFCN3K} (close to $1.5$M if we add all the ILSVRC-Detection annotations). Similarly, OpenImages has more than $8$K trainable categories but only $601$ categories have bounding boxes~\cite{openimages}. 

Although researchers have developed several large scale algorithms, such as R-FCN 3000~\cite{RFCN3K} and SNIPER with soft sampling~\cite{Wu2018, Singh2018} to facilitate detection on the scale of ImageNet and OpenImages \emph{detection set}. These methods still require full annotations of ground truth bounding boxes. This not only wastes a large part of the training data, which could potentially help learning a more robust detector, but also limits the detector's capability to recognize more fine-grained categories. In this sense, fully-supervised detectors handicap our capability to utilize all training data and detect more categories. 
We believe a better way to train large-scale fine-grained detector is through semi-supervised learning with coarse-grained detection data and fine-grained classification data. There are several benefits for such semi-supervised detector. First, compared to fully-supervised detectors, the proposed semi-supervised detector only needs a relatively small amount of coarse-grained classes to be fully annotated. Intuitively, if we have fully annotated ``dog'', we do not need bounding boxes for all the species of dogs.  This greatly reduces the annotation cost and makes better use of existing data. Secondly, compared to weakly-supervised detectors, we can exploit the coarse-grained detection data and train a stronger detector. State-of-the-art weakly-supervised methods are still $30$ points below fully-supervised methods~\cite{Ge2018, Zhang2018, Tao2018} in mAP,  while our semi-supervised method demonstrates that we can achieve comparable performance to fully-supervised methods. 

The key to solve the problem, in our opinion, lies in the correlations between coarse-grained and fine-grained classes. Thus we are trying to answer the following two questions: 1) How to build proper correlations between fully-supervised coarse-grained data and weakly-supervised fine-grained data. 2) How to effectively utilize these correlations to transfer accurate object appearances learned from fully-supervised to weakly-supervised data, as well as to learn a better detector from the rich variances of more fine-grained weakly-supervised data.

To answer these questions, we propose an innovated large scale semi-supervised object detection solution. Compared to existing works~\cite{Redmon2017, RFCN3K}, our solution is able to handle both semantic and visual correlations between fully-supervised and weakly-supervised set, which makes the solution more extendable and applicable for more tasks. More importantly, beyond shared feature learning layers, we explicitly exploit these correlations to transfer knowledge from both data worlds. This not only helps us extend the detection to fine-grained level, but also improve the detector performance on coarse-grained labels with the help of diverse fine-grained data. 

Specifically, we propose a two-stream network with a backbone based on R-FCN. One stream focuses on fully-supervised detection and the other stream solves the weakly-supervised detection with fine-grained data. These two streams share feature learning layers. Our main technical contributions include:
\squishlist
    \item Fully-supervised detection stream augmentation with the fine-grained weakly-annotated data. With the shared backbone and multi-task training, we make use of more diverse fine-grained images with weak annotations to boost the detector performance.
    \item Soft-attention based proposal re-ranking for weakly-supervised detection stream. We utilize the correlations between coarse- and fine-grained labels to bring attentions to proposals that are more likely to contain related objects from fully-supervised stream to the weakly-supervised stream.
    \item Dual-level memory module with foreground attention pooling. We augment the network with an external memory module. Similar to clustering loss, this module can transfer knowledge from supervised to unsupervised data and further regularize the training process.
\squishend

Results on two large scale datasets - OpenImages and ImageNet demonstrate that our method, without the need of bounding boxes annotations, can still achieve high detection accuracy on fine-grained classes, while also boosting the detection performance on coarse-grained classes in some cases. Furthermore, our designed framework is end-to-end trainable and almost as efficient as a standard detection network. The proposed components are well generalized and can be easily transferred to any 2-stage (RPN based) detectors in order to keep up with the rapid development of deep learning. The success of our method provides meaningful insights for practical detection data collection: to build a large scale object detector that is capable of detecting tens of thousands of classes (e.g. the 11K classes in ImageNet dataset), what we only need is to collect bounding boxes for a small set of coarse-grained classes and image labels for all fine-grained classes, which can significantly reduce the annotation cost.

\vspace{-0.05in}
\section{Related Work}

\textbf{Fully-Supervised Object Detection:} Fully-supervised detection can be divided into two categories: 1) One-stage methods, such as YOLO series~\cite{Redmon2016, Redmon2017, Redmon2018} and SSD~\cite{Liu2016}. These methods do not require regional proposals and perform detection in a single shot. 2) Two-stage methods, such as Fast R-CNN series~\cite{Girshick2015, Ren2017}, R-FCN~\cite{Dai2016}, FPN~\cite{Lin2017} and Mask-RCNN~\cite{He2017}. These methods build on the same idea that a detector should first generate regional proposals, which are likely to contain objects, then further classify the proposals into background and specific object classes. One-stage methods in general are faster in training and inference, but with lower detection accuracy, compared to two-stage methods. Our proposed framework is built on top of one of the state-of-the-art method, R-FCN with Deformable Convolutional Network~\cite{Dai2016, Dai2017}. However, it can be easily adopted to any other two-stage detectors.

\textbf{Weakly-Supervised Object Detection:} Weakly-supervised object detection is often formulated as the key instance detection on multi-instance learning, where we view each object proposal as an instance and each image as a bag. The problem is to find out instances that contain objects, given only bag-level supervision. Most weakly-supervised detection methods have two stages: they first utilize Selective Search~\cite{Uijlings2013} or Edge Boxes~\cite{Zitnick2014} as proposals, and then use a CNN to solve the multi-instance learning problem. There are two main directions to further improve the performance of weakly supervised detection: improving the proposal quality and the aggregation-selection process of proposals. WSDDN~\cite{Bilen2016} is one of the well-known work for weakly-supervised object detection with deep learning. The key idea is to have an additional ranking softmax for proposals to smartly aggregate and select proposal scores. OICR~\cite{Tang2017} improves WSDDN by incorporating multiple refinement streams with pseudo ground-truth. ~\cite{Tao2017} exploits web images to enhance training data. ~\cite{Tang2018} ditches hand-crafted object proposals in favor of a weakly-supervised version of Regional Proposal Network (RPN). Recently, ~\cite{Ge2018} also proposes to use extra attention map to improve the proposal selection process. In this work, we show that bounding box information of correlated coarse-grained detection classes can drastically improve the weakly-supervised fine-grained stream in both directions.

\textbf{Semi-Supervised Object Detection:} There are only a few research works in the semi-supervised detection field. ~\cite{Tang2016} proposes a LSDA-based method that can handle disjoint set semi-supervised detection, but this method is not end-to-end trainable and cannot be easily extended to state-of-the-art detection frameworks. ~\cite{Uijlings2018} proposes a semiMIL method on disjoint set semi-supervised detection, which achieves better performance than~\cite{Tang2016}. Note-RCNN~\cite{Gao2018} proposes a mining and training scheme for semi-supervised detection, but it needs seed boxes for all categories.  YOLO 9000~\cite{Redmon2017} can also be viewed as a semi-supervised detection framework, but it is no more than a naive combination of detection and classification stream and only relies on the implicit shared feature learning from the network. 

To the best of our knowledge, our method is the first semi-supervised fine-grained detection framework that explicitly exploit semantic/visual correlations between coarse-grained detection and fine-grained classification data. Experimental results on ImageNet and OpenImages datasets demonstrate that our setting is more applicable for real world large scale detection problems.

\section{Technical Approach}

In this section, we introduce how we solve the problem of semi-supervised fine-grained detection. We first formulate the problem in Section.~\ref{formulation}, then introduce how we encode visual and semantic correlations in Section.\ref{correlations}. Our overall architecture is outlined in Section.\ref{architecture}. The key components are: a fully-supervised detection stream, a weakly-supervised stream with soft-attention based proposal re-ranking and a dual-level memory module with foreground attention pooling, which are explained in detail in Section.\ref{fully}, Section.\ref{weakly} and Section.\ref{memory}, respectively. 

\subsection{Problem Formulation}
\label{formulation}

Let $\mathcal{X}$ be the whole dataset consisting of a subset $\mathcal{X}_f$ of $\mathcal{C}_f$ classes with full ground-truth bounding box annotations ($f$ denotes for fully-annotated), and a subset $\mathcal{X}_w$ of $\mathcal{C}_w$ classes with only image-level annotations ($w$ denotes for weakly-annotated). Let $|\mathcal{C}_f| = C_f$ and $|\mathcal{C}_w| = C_w$. Our goal is to train a detector that is able to accurately detect all $\mathcal{C} = \mathcal{C}_f \cup \mathcal{C}_w$ classes. Our methods target to handle the 
challenging scenario, where the fully-annotated and weakly-annotated classes are disjoint (i.e., $\mathcal{C}_f \cap \mathcal{C}_w = \varnothing$), and there are much more image level annotations than bounding box annotations (i.e., $C_w \gg C_f$). This scenario includes the most prominent large-scale image datasets, namely ImageNet and OpenImages.

We assume that there exists semantic or/and visual correlations between fully-annotated set $\mathcal{C}_f$ and weakly-annotated set $\mathcal{C}_w$. We also assume that $\mathcal{C}_f$ contains all {\it coarse-grained} level labels and $\mathcal{C}_w$ contains all {\it fine-grained} level labels to reflect real world scenarios. For example, for semantic correlation, we could have ``dog'' as a coarse-grained label with bounding boxes, and ``labrador'', ``chihuahua'', etc as fine-grained labels without bounding box annotations. Or for visual-semantic correlation, we could have ``dog'' as a coarse-grained label, and ``wolf'', ``coyote'', or even ``stuffed dog'' as fine-grained labels. Our method aims to utilize these correlations to provide accurate bounding boxes predictions for fine-grained classes and improve the training accuracy on coarse-grained classes with related fine-grained data.

\subsection{Encoding Correlations between Coarse-Grained and Fine-Grained Classes}
\label{correlations}
We believe the key of a successful semi-supervised detector is to build and utilize correlations between coarse-grained and fine-grained classes. These correlations are bridges that allow us to transfer knowledge between two worlds. Specifically, we consider two kinds of correlations: semantic and visual correlations.
\vspace{-0.1in}
\subsubsection{Semantic Correlations}
Semantic correlations are extracted from human knowledge and languages on how objects/concepts are structured and related. These correlations are often represented as a directed graph or tree, such as WordNet~\cite{Miller1995} and Visual Genome~\cite{Krishna2016}. The benefits of such semantic correlations are: 1) They encode strong prior knowledge on how we view the world. 2) They are readily available from various sources. For example, ImageNet is built on WordNet, and OpenImages is built on Google Knowledge Graph. 

Encoding semantic correlations are straightforward as these correlations are already represented by a directed graph. For a coarse-grained detection class $c_i^f$, if we consider its hyponyms as $\mathcal{H}(c_i^f)$, the encoding function can be written as a one-hot vector:
\begin{equation}
\label{semantic-encoding}
    \mathcal{M}(c_i^f) = [e_1^w, \dots , e_{C_w}^w],  
    \text{     where    } 
\begin{cases}
    e_j^w = 1,& \text{if } c_j^w \in \mathcal{H}(c_i^f)\\
    0,              & \text{otherwise}.
\end{cases}
\end{equation}
Here $c_j^w$ is an arbitrary fine-grained classification class. The reverse encoding from fine-grained to coarse-grained is just a similar function with hypernyms. We use these mapping functions in subsequent experiments.

There are drawbacks on semantic correlations. Firstly, adding new nodes to the existing graph requires a lot of expert efforts and is prune to errors. Secondly, semantic correlations are not always transferable to visual similarities, and vice versa. For example, ``hyena'' biologically is closer to ``feline'', but visually more similar to ``canine'', and ``basketball'' is visually similar to ``orange'', but semantically they are far away. Therefore, we introduce visual correlations in the next section.
\vspace{-0.1in}
\subsubsection{Visual Correlations}
Visual correlations describe the visual similarities among objects/concepts. These correlations align better with training objectives, and are more flexible and easier to maintain. To encode visual correlations, we consider two scenarios, directly using detection classes as ``super-classes'' or building ``super-classes'' by clustering.

If we have relatively small number of coarse-grained classes, we could treat each of them as a ``super-class'' and build a two-level encoding. We obtain the $i$-th object-class representation, $x_i$ , by taking the average of features $x_{i,j}$ (extracted from the final layer of a deep neural network, such as ResNet-101, for sampled images $j$ belonging to the $i$-th class). After acquiring the representation for each class, we can then encode the correlations between coarse-grained and fine-grained classes through either hard-assignment or soft-assignment. 

Let's denote $\scriptstyle d_i^j = \left\|x_i^f-x_j^w\right\|_2$ as the Euclidean distance between two representations $x_i^f$ from coarse-grained set and $x_j^w$ from fine-grained set, the hard-assignment encoding function is similar to Eq.~\ref{semantic-encoding}:
\begin{equation}
\label{visual-encoding-hard}
    \mathcal{M}(c_i^f) = [e_1^w, \dots , e_{C_w}^w],
    \text{        where }
\begin{cases}
    e_j^w = 1,& \text{if } d_i^j < \theta_i\\
    0,              & \text{otherwise}.
\end{cases}
\end{equation}
Here $\theta_i$ is the class-specific threshold.

For soft-assignment, which is similar to weighted K-means clustering, we can assign a fine-grained class to multiple/all coarse-grained classes using a softmax function:
\begin{equation}
\label{visual-encoding-soft}
    \mathcal{M}(c_i^f) = [e_1^w, \dots , e_{C_w}^w],
    \text{   where  }
    e_j^w = \frac{e^{-\beta d_i^j}}{\sum_k{e^{-\beta d_k^j}}},
\end{equation}
where $\beta$ is the temperature parameter that control the distribution of softmax function.

If we have a large number of coarse-grained classes, to reduce computational cost as well as facilitate more effective visual encoding, we can build a set of super-classes $\mathcal{C}_s$ by (weighted, if we use soft assignment encoding) K-mean clustering on the representations of object classes from the coarse-grained set $\mathcal{C}_f$. We then have two encoding functions for $\mathcal{C}_s \rightarrow \mathcal{C}_f$ and $\mathcal{C}_s \rightarrow \mathcal{C}_w$ in the same spirit of  Eq.~\ref{visual-encoding-hard} and Eq.~\ref{visual-encoding-soft}. 

\subsection{Architecture}
\label{architecture}

\begin{figure}
\centerline{\includegraphics[width=.5\textwidth]{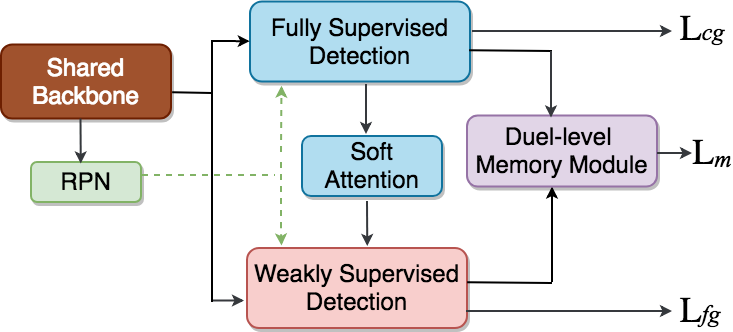}}
\caption{The overview of the designed architecture. The architecture can be split into three streams: 1) a {\it fully-supervised detection stream}, 2) a {\it weakly-supervised classification stream}, and 3) the {\it Correlation components including the soft-attention based proposal re-ranking and a dual-level memory module}. All three streams share common modules such as base CNN layers for feature learning and regional proposal network (RPN). During training, the detection data are used to train the RPN and R-FCN alike coarse-grained detection stream, while the fine-grained data are used to train the fine-grained classification stream. The correlation components are designed to transfer knowledge between coarse-grain and fine-grain.}
\label{fig:full}
\end{figure}

To utilize the fully-supervised (coarse-grained) and weakly-supervised (fine-grained) data and their encoded correlations, we build three key components in our framework: 1) a fully-supervised detection stream for coarse-grained classes, 2) a weakly-supervised classification stream for fine-grained classes, and 3) correlation components to transfer knowledge between coarse-grained and fine-grained data，including shared backbone, soft-attention based proposal re-ranking, and the dual-level memory module. Details of these components are explained in the following sections.

\vspace{-0.2in}
\subsubsection{Fully-Supervised Detection Stream}
\label{fully}

This stream is built on Deformable R-FCN~\cite{Dai2016}. First, a Regional Proposal Network (RPN) is used for generating proposals and is only trained with detection data to avoid label noises from weakly supervised data. Then, on the shared backbone, we apply position sensitive filters to pool features from each proposal. Since we have $C_f$ classes and $P \times P$ filters per class, there are $(C_f+1)\times P\times P$ filters. After performing position-sensitive RoI pooling, we apply two fully connected layers to obtain final classification scores and regress bounding box results for each proposal. A cross-entropy loss and a bounding box regression loss are used for classification and regression learning, respectively:
\begin{equation}
   \mathcal{L}_{cg} = 0.5 * \mathcal{L}^{reg}_{cg} + \mathcal{L}^{cls}_{cg},
\end{equation}
where we use $0.5$ as the trade-off parameter~\cite{Singh2018}.

\subsubsection{Weakly-Supervised Detection Stream with Soft-Attention Proposal Re-ranking}
\label{weakly}
\begin{figure}[h]
\centerline{\includegraphics[width=.5\textwidth]{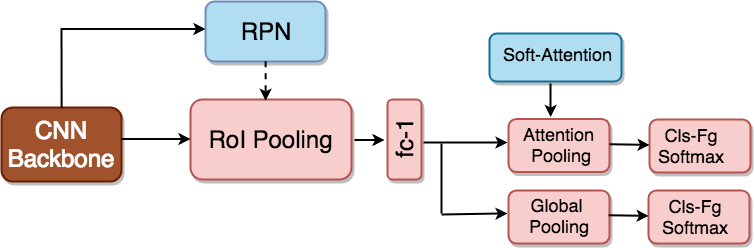}}
\caption{The overview of weakly-supervised stream.}
\label{fig:fg}
\end{figure}

Weakly-supervised detection can be viewed as a special kind of multi-instance multi-label learning with key instance detection. Each image can be a bag in multi-instance learning and each proposal from the image is an instance in the bag. Since we just have the image(bag)-level labels, the key for solving this problem is how to aggregate proposal-level scores into image-level scores and how we select proposals that are most likely to contain target objects. In the context of deep learning and back-propagation, these two problems are closely tied together.

A classical way to aggregate proposal scores is to use max or average pooling~\cite{Yang2017}. This works well for getting good image-level predictions, but not for detecting proposals with the highest Intersection over Union (IoU) to ground truth bounding boxes. The most likely reason is that modern CNNs tend to focus on the most discriminative part of objects rather than the whole object~\cite{Zhang2018}. For example, networks trained on ImageNet differentiate ``person'' from other objects by only using the head/face part rather than the whole body. Therefore, state-of-the-art weakly-supervised object detection methods~\cite{Bilen2016, Tao2017, Tao2018, Ge2018} employ some forms of regularization on the activation maps or the proposals selection to solve the problem.

Fortunately, since we have the fully-supervised detection stream, for each proposal, we actually know the presence or absence of closely related coarse-grained objects for a fine-grained object. Continuing with our ``dog'' example, considering if we are training the model for ``chihuahua'', a proposal with high score on ``dog'' from the fully-supervised detection stream is much more likely to contain ``chihuahua'' than a proposal with low score on ``dog''. This knowledge from the detection data is one good attention mechanism for weakly-supervised detection.

We design our weakly-supervised branch based on this idea. Similar to the fully-supervised stream, we use the shared RPN to generate proposals, and a RoI pooling layer to exact features for each proposal from shared layers. We select RoI pooling instead of PSRoI pooling to reduce the computation overhead, as PSRoI pooling requires $P \times P$ times more filters than RoI pooling, and there are a large number of classes in our case. After we generate pooled features, we obtain proposal-level scores through a fully connected layer. These scores are then sent to two different pooling branches as shown in Fig.\ref{fig:fg}. The first branch is the global pooling: we use max, average or weighted average pooling to aggregate image-level scores, and normalized softmax loss to learn. This branch is designed for smoothing the training process and generating good image-level performance. The second branch is the attention pooling, which transfers knowledge from fully-supervised stream and helps create accurate proposals. Similar to soft-attention mechanism in neural machine translation~\cite{Bahdanau2014}, we use proposal scores from fully-supervised detection stream as an attention map and apply this map to the weakly-supervised detection scores to obtain final proposal scores. This score is then aggregated by average pooling.

\begin{figure}
\centerline{\includegraphics[width=.45\textwidth]{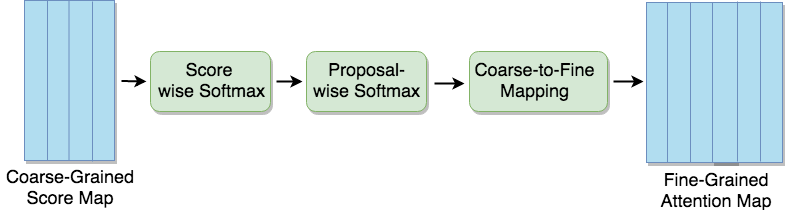}}
\caption{Soft-attention based proposal re-ranking.}
\label{fig:attention}
\end{figure}

As shown in Fig.\ref{fig:attention}, to obtain the attention map for ranking, we need to re-scale and normalize the coarse-grained detection scores, and then map the scores to fine-grained labels using previously discussed mapping functions, i.e., Eq. (\ref{semantic-encoding}), (\ref{visual-encoding-hard}), (\ref{visual-encoding-soft}). Assuming we have a score map $\mathcal{S}^f \in \mathbb{R}^{(C_f+1) \times P}$, where $C_f+1$ is the number of coarse-grained detection classes plus background, and $P$ is the number of proposals. If background class is at index $C_f+1$, to constrain that each proposal on the map to one unique class, we use a softmax operation defined as:
\begin{equation}
    \hat{\mathcal{S}}^f(p, c) = \frac{e^{\mathcal{S}^f(p, c)}}{\sum_{j=1}^{C_f}{e^{\mathcal{S}^f(p, j)}}}\text{,   }
    \forall c \in \left[1, \dots,  C_f\right], 
\end{equation}
where $\mathcal{S}^f(p, c)$ is the score for proposal $p$ at class $c$. After obtaining $\hat{\mathcal{S}_c^f}$, similar to the ranking term in WSDDN~\cite{Bilen2016}, we normalize the score map in the direction of proposals to obtain the coarse-grained attention map
\begin{equation}
   \mathcal{A}^f(p, c) =\frac{e^{\hat{\mathcal{S}}^f(p, c)}}{\sum_{j=1}^{P}{e^{\hat{\mathcal{S}}^f(j, c)}}} \text{,   }
    \forall p \in \left[1, \dots,  P\right], 
\end{equation}
However, since we need to apply the attention map to fine-grained proposals, we utilize the encoding function described in Section~\ref{correlations} as a coarse-to-fine mapping function to map the coarse-grained attention map to fine-grained. We obtain the fine-grained attention map $\mathcal{W}^w(p)$ at proposal $p$ as
\begin{equation}
    \mathcal{A}^w(p) = \sum_{j=1}^{C_f} \mathcal{A}^f(p, j) * \mathcal{M}(j),
\end{equation}
where $\mathcal{M}(j)$ is the soft-assignment or one-hot hard-assignment encoding on class $j$, with dimension $C_w$. This attention map is then applied to the weakly-supervised score map $\mathcal{S}^w \in \mathbb{R}^{C_w \times P}$ by element-wise product. 

The final loss function of the weakly-supervised fine-grained detection branch is:
\begin{equation}
   \mathcal{L}_{fg} = \mathcal{L}^{cls}(\text{pool}(\mathcal{S}^w), y) + \lambda \mathcal{L}^{cls}(\text{pool}(\mathcal{S}^w \odot \mathcal{A}^w), y),
\end{equation}
where $y$ is a multi-label image level label and $\lambda$ is the trade-off parameter which is set to be $0.1$ for all of our experiments. We use Top-$5$ average pooling for the first term and sum pooling for the second term. Detailed experiments on design choices can be found in our supplemental file.

\vspace{-0.1in}
\subsubsection{Dual-Level Memory Module}
\label{memory}
Neural networks with memory are recently introduced to enable more powerful learning and reasoning ability for addressing several challenging tasks, such as question answering\cite{Miller2016}, one-shot learning\cite{Kaiser2017} and semi-supervised classification~\cite{Chen2018}. Augmenting a network with an external memory component provides a similar role as the clustering loss in standard semi-supervised learning, but with dynamically updating feature representations and probabilistic predictions memorization.

Inspired by these works, we propose to add a memory module to our framework to take advantage of the memorable information generated in model learning and to further regularize the learning. Unlike~\cite{Chen2018}, our semi-supervised fine-grained detection task is a two-level semi-supervised problem. For fully-supervised detection stream, we do not have proposal-level coarse-grained labels for the fine-grained classification data, and for weakly-supervised stream, we do not have image-level fine-grained labels for the coarse-grained detection data. Therefore, we need two levels of memory, coarse-grained proposal-level memory and fine-grained image-level memory. With these principles in mind, we propose a Dual-Level Memory module with Foreground Attention pooling (DLM-FA).

For the coarse-grained level memory, we have proposal(box)-level labels for coarse-grained detection images, but we are lacking proposal-level coarse-grained labels for fine-grained images. If we view each proposal as one training instance, we are facing a straightforward semi-supervised learning problem and we can directly use the same memory structure as in~\cite{Chen2018}. The loss function is:
\begin{equation}
   \mathcal{L}_m^w = H(\hat{\vec{p}}) + D_{KL}(\vec{p}||\hat{\vec{p}}),
\end{equation}
where $H(\cdot)$ is the entropy and $D_{KL}(\cdot)$ is the Kullback-Leibler (KL) divergence. $\hat{\vec{p}}$ is the memory prediction and $\vec{p}$ is the network prediction for each proposal $p$.

For the fine-grained level memory, we have image-level labels for fine-grained classification images, but we are lacking fine-grained labels for coarse-grained images. However, compared with standard semi-supervised setting, we are dealing with multi-instance semi-supervised scenario. Therefore, we need to aggregate proposal-level features and predictions to image-level in order to facilitate memory update and prediction. We use Foreground Attention (FA) pooling to filter out noisy proposals. In FA pooling, we only pool features and predictions from proposals with high responses of positive image-level, and aggregate these features and predictions by sum pooling to represent their corresponding images. Specifically, if image $I$ has $m$ proposals ${p_i}_{i=1}^m$, and their corresponding features and scores are $\{p_i^f\}_{i=1}^m$ and $\{p_i^s\}_{i=1}^m$, the image-level feature $I^f_c$  for a specific class $c$ is then defined by:
\begin{equation}
   I^f_c = \sum_i(p_i^f),  \text{ if } \text{argmax}(p_i^s) = c,
\end{equation}
The image-level predictions $I^s_c$ is pooled in a similar way. After we have the image-level features and scores to update memory modules, we utilize
\begin{equation}
   \mathcal{L}_m^f = H(\hat{I}) + D_{KL}(I||\hat{I}),
\end{equation}
as the loss function for fine-grained level memory. The whole memory loss is then:
\begin{equation}
   \mathcal{L}_m = \mathcal{L}_m^w + \mathcal{L}_m^f
\end{equation}

Details of the memory module can be found in the supplementary file.

\section{Experiments}

\label{exp}
In this section, we first introduce our implementation details. Then we discuss the experimental results and compare them with other baseline methods. We test our method on two most challenging large scale datasets -- OpenImages~\cite{openimages} and ImageNet~\cite{Deng2009}. 

\vspace{-0.04in}
\subsection{Implementation Details and Baselines}
\label{subset:impl}
Our implementation is based on SNIPER~\cite{Singh2018}. In particular, we use mixed precision training for larger batch size and faster training speed. ResNet-$101$ with fp$16$ weights is used as the shared backbone. We use fp$32$ weights for the fully connected and convolutional layers of all heads. We train the model on $8$ V-$100$ GPUs with a batch size of $128$ (i.e. $16$ per GPU). A balance sampling scheme is used for detection and classification data, i.e., we sample the same number of classification and detection data for each batch. The initial learning rate is set to be $0.015$ for all experiments. We train all models for $9$ epochs and the learning rate is dropped by $0.1$ for every $3$ epochs. Image horizontal flipping is used for data augmentation. We only use one scale, namely $512 \times 512$ for both training and testing. During inference, we run soft NMS~\cite{Bodla2017} on the model outputs with the standard deviation parameter of $0.55$ in the Gaussian weighting function.

We compare our method with state-of-the-art fully-supervised detection methods trained on the same data. All results are reported in mean average precision (mAP) with intersection-over-union (IoU) threshold at $0.5$. Specifically, our method is compared to: 1) {\bf SNIPER-CG-Fully}: SNIPER trained on coarse-grained data with fully annotated bounding boxes. 2) {\bf SNIPER-FG-Fully}, i.e., SNIPER trained on fine-grained data with fully annotated bounding boxes. 3) {\bf SNIPER-FG-Weakly}: SNIPER with fixed backbone and RPN trained from coarse-grained detection data and fine-tune on fine-grained data with only image-level labels. This is one of the strongest weakly baselines in our test. 4) {\bf SNIPER-All}: SNIPER trained on coarse- and fine- grained data with all bound box and label annotations. 
\footnote{We have also compared with semi-supervised detection methods ~\cite{DOCK, Tang2016, Uijlings2018} in same-granularity random-split setting. Since this is not the main focus of our paper and due to space limitations, the results are shown in supplemental file.} 
\setlength{\textfloatsep}{0.35cm}

\begin{figure*}[h]
    \centering
    \footnotesize
    \subfloat{\includegraphics[width=0.2\textwidth]{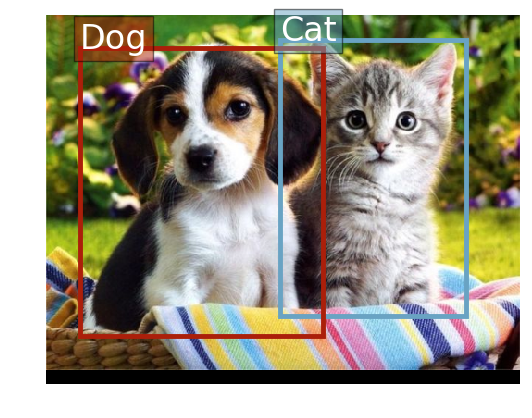}}\hfill
    \subfloat{\includegraphics[width=0.2\textwidth]{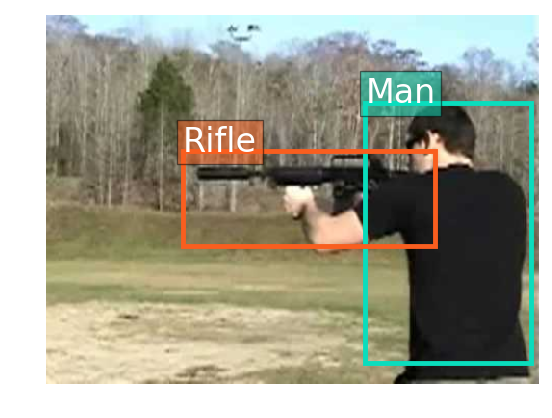}}\hfill
    \subfloat{\includegraphics[width=0.18\textwidth]{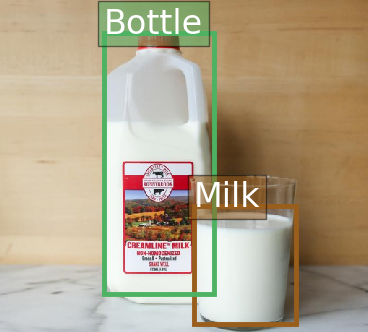}}\hfill
    \subfloat{\includegraphics[width=0.2\textwidth]{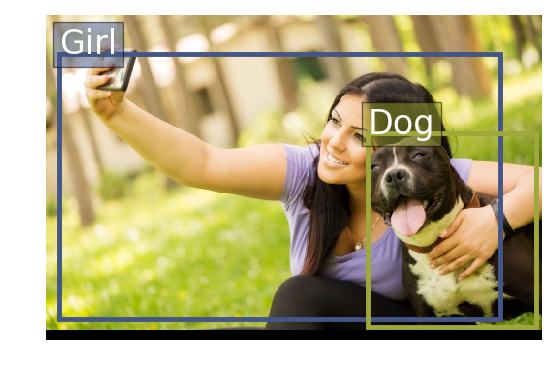}}\hfill
    \subfloat{\includegraphics[width=0.2\textwidth]{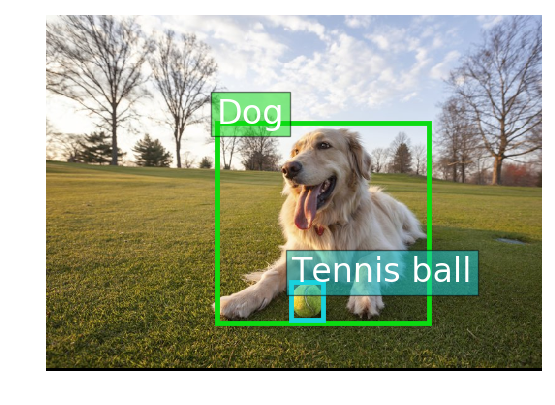}}
        
    \subfloat{\includegraphics[width=0.19\textwidth]{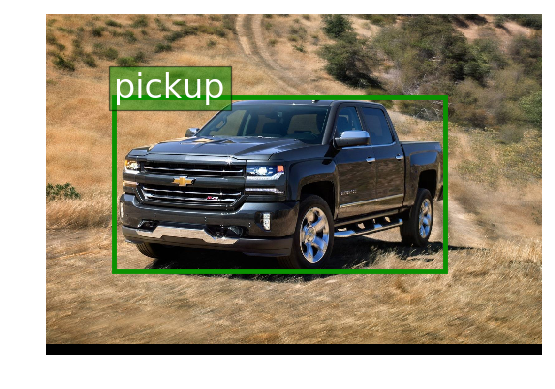}}\hfill
    \subfloat{\includegraphics[width=0.18\textwidth]{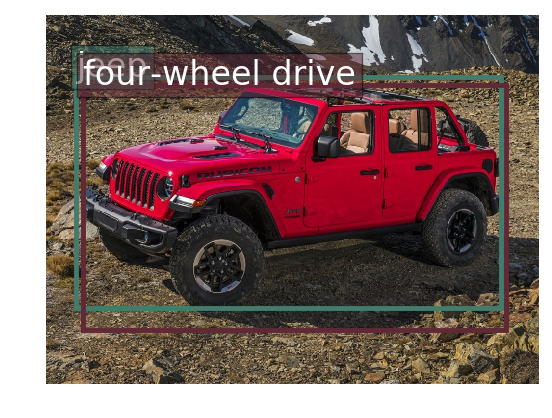}}\hfill
    \subfloat{\includegraphics[width=0.20\textwidth]{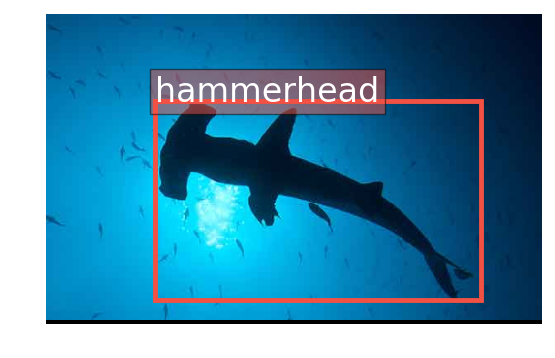}}\hfill
    \subfloat{\includegraphics[width=0.19\textwidth]{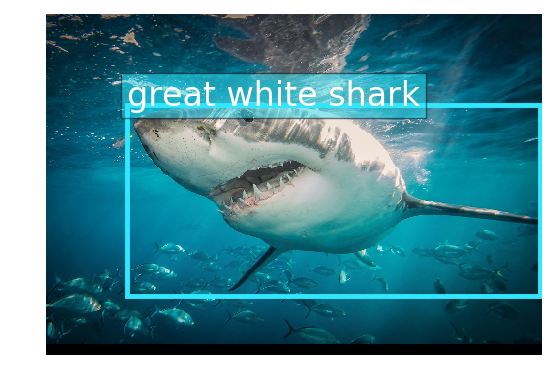}}\hfill
    \subfloat{\includegraphics[width=0.19\textwidth]{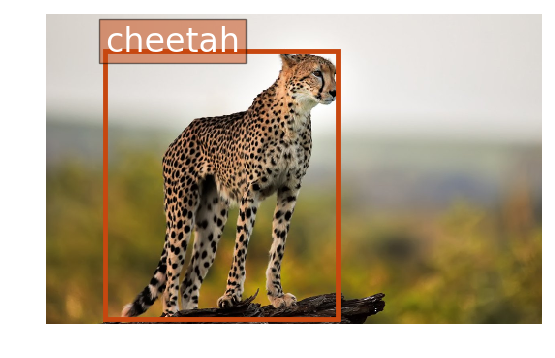}}%
    \caption{Qualitative Results. The first row shows the fine-grained detection results from our best OpenImages model (last row of Table~\ref{tab:oi_results}). The second row shows fine-grained detection results from our ImageNet model on $11K$ fine-grained classes.}
    \label{fig:qualitative}
\end{figure*}

\vspace{-0.04in}
\subsection{OpenImages Results}
\begin{table}[h]
    \centering
    \footnotesize
    \begin{tabular}{c|c|c|c}
    \hline
    Class & $\#$ classes & $\#$ training & $\#$ test  \\
    \hline
    Coarse-grained & 34 & 786K & 22K \\
    \hline
    Fine-grained & 462 & 567K & 8.8K \\
    \hline
    \end{tabular}
    \caption{Statistics for OpenImages dataset.}
    \label{tab:open_data}
\end{table}

OpenImagesV4 dataset contains bounding box annotations for $601$ classes with a semantic tree based on Google Knowledge Graph. We use all the $462$ leaf nodes in the semantic tree as fine-grained classes. These fine-grained classes have $72$ direct parent-nodes that can be used as coarse-grained classes. However, since these $72$ classes also have hierarchy, for simplicity, we merged the lower level classes into their highest level parent classes. We end up with $34$ coarse-grained classes. 
As discussed in the previous section, we only use bounding box annotations of the coarse-grained classes, and image-level labels for fine-grained classes in training. We evaluate our model on the OpenImages validation dataset for object detection. The validation dataset contains bounding box annotations for all coarse-grained and fine-grained labels. Table~\ref{tab:open_data} shows the statistics for the coarse-grained and fine-grained data in training and validation datasets of OpenImages.

Results for different models are sumarrized in Table~\ref{tab:oi_results}. We can see that naive combination of the fully-supervised and weakly-supervised stream without soft-attention and memory can already perform pretty well on both tasks, especially on coarse-grained detection. The naive baseline already outperforms SNIPER-CG-Fully by around $7$ points. This implies that the shared backbone benefits a lot from the rich variety of the \emph{closely related} fine-grained images, as well as the multi-task training. However, due to the simple proposal aggregation method (top-5 pooling), we can see that the fine-grained detection results, though reasonable, are far lower than SNIPER-FG-Fully and SNIPER-All. We can also see that naive weakly-supervised stream still outperforms the pure weakly-supervised method on the same RPN, which further validates the benefits of joint training and shared backbone.

If we add soft-attention based proposal re-ranking during training and inference, the $mAP$ on fine-grained class increases by over $15$ points. It is clear shown that, bringing the knowledge learned from the detector in can significantly help the weakly-supervised stream. For the memory module, if we just add a single-level coarse-grained memory module similar to~\cite{Chen2018}, we do not see any improvement. However, with the help of Dual-Level Memory module, we can further minimize the gap between our fine-grained stream and the fully-supervised one.

\begin{table}[t]
    \centering
    \footnotesize
    \begin{tabular}{c|c|c|c}
        \hline
        Method & Training Data/Label  & mAP-CG & mAP-FG\\
        \hline
        SNIPER~\cite{Singh2018} & CG-Fully  & 45.7 &-  \\
        \hline
        SNIPER~\cite{Singh2018} & FG-Fully  & - & \bf{59.1} \\
        \hline
        SNIPER~\cite{Singh2018} & All  & 28.7 & 54.0 \\
        \hline
        SNIPER-Weakly & FG-Weakly  & -  & 20.2 \\
        \hline
        \hline
        Naive & CG-Fully+FG-Weakly  & 52.5  & 34.0 \\
        \hline
        +Soft-Attention & CG-Fully+FG-Weakly &52.9  & 49.2  \\
        \hline
        +CG-Memory & CG-Fully+FG-Weakly &52.2  & 49.6  \\
        \hline
        +DLM-FA & CG-Fully+FG-Weakly &\bf{53.5} & 51.9   \\
        \hline
    \end{tabular}
    \caption{Test results for different models on OpenImages dataset where we evaluated coarse-grained (CG) and fine-grained (FG) classes separately.}
    \label{tab:oi_results}
\end{table}

Although it shows that our best model is still $7$ points lower than the best fully-supervised model on fine-grained classes, we should mention that, our model actually performs similarly with SNIPER-FG-Fully in terms of $mAP_{0.5:0.95}$, both at around $36$. This demonstrates that though we do not have bounding boxes on fine-grained classes, we can still learn to predict as accurately as fully-supervised methods.

\vspace{-0.04in}
\subsection{ImageNet Results}
\vspace{-0.04in}
\begin{table}[h]
    \centering
    \footnotesize
    \begin{tabular}{c|c|c|c}
    \hline
    Class & $\#$ classes & $\#$ training & $\#$ test  \\
    \hline
    Coarse-grained & 200 & 400K & 22K \\
    \hline
    Fine-grained-3K & 2937 & 870K & 46K \\
    \hline
    Fine-grained-11K & 11021 & 1.7M & - \\
    \hline
    \end{tabular}
    \caption{Statistics for ImageNet dataset.}
    \label{tab:imagenet_data}
\end{table}

\begin{table}[ht]
    \centering
    \footnotesize
    \begin{tabular}{c|c|c|c}
        \hline
        Method & Training Data/Label  & mAP-CG & mAP-FG \\
        \hline
        SNIPER~\cite{Singh2018} & CG-Fully  & \bf{54.0} & - \\
        \hline
        SNIPER~\cite{Singh2018} & $3$k-FG-Fully  & - & \bf{41.6} \\
        \hline
        YOLO-$9000^*$~\cite{Redmon2017} & COCO+$9k$-FG-Weakly  & 19.9 & - \\
        \hline
        R-FCN-$3000^*$~\cite{RFCN3K} & $3$k-FG-Fully  & 34.9 & - \\
        \hline
        \hline
        Ours-3K & CG-Fully+$3$k-FG-Weakly  & 50.7  & 35.1  \\
        \hline
        Ours-11K & CG-Fully+$11$k-FG-Weakly  & 49.1  & -  \\
        \hline
    \end{tabular}
    \caption{Test results for different models on ImageNet dataset where we evaluated coarse-grained (CG) and fine-grained (FG) classes separately. Please note that YOLO-$9000$ and R-FCN-$3000$ results are not directly comparable. Based on~\cite{RFCN3K}, SNIPER-$3k$-FG-Fully should be the performance upper bound of R-FCN-$3000$ on FG.}
    \label{tab:in_results}
    \end{table}
\vspace{-0.1in}
 We then run experiments on ImageNet dataset. As shown in Table~\ref{tab:imagenet_data}, we use the ILSVRC 2014 Detection set with $200$ classes as the coarse-grained set. Two fine-grained sets are tested. One is the $3K$ set with bounding box annotations, similar to what is used in~\cite{RFCN3K}. For this set, we split $5\%$ of the training data to the validation set to test the performance of our fine-grained stream. The other set contains $11$K classes with each class having over $500$ training images. There are in total 13M images in this set. In our experiment, we randomly sample $1/8$ images from all classes for training and testing. Such a subset could have been general enough and is a good representative of the full set. We build visual correlations between the fine-grained and coarse-grained classes using the soft assignment introduced before.

From the results summarized in Table~\ref{tab:in_results}, we can see that unlike results from OpenImages dataset, our coarse-grained detection result is slightly worse than SNIPER-CG-Fully, by margins of $3.3$ and $4.9$, respectively. This could be explained by the correlations of coarse-grained and fine-grained data. For OpenImages, the coarse-grained and fine-grained sets are much closely correlated as they are hand-picked to form a compact semantic tree, while for ImageNet, the fine-grained sets are arbitrary picked by the availability of bounding box annotations and the number of training images. The ImageNet dataset could contain much more diverse classes in terms of semantic and visual correlations between the coarse-grained set. Therefore, we do not observe improvement on the coarse-grained performance. 

Similar to OpenImages, though our best model is $6$ points worse than the best supervised model on fine-grained classes in $mAP_{0.5}$, our model actually performs better than SNIPER-FG-Fully in terms of $mAP_{0.5:0.95}$, with our model at $25$ and SNIPER-FG-Fully at $22$. This again demonstrates that we can learn to predict as accurately as fully-supervised methods on large scale datasets.

We also present the detection results of YOLO-$9000$ and R-FCN-$3000$ on the CG validation set in Table~\ref{tab:in_results}. Note that these methods are not trained on the same data as our method, thus the results cannot be directly compared. However, what we would like to present here is that we are able to train a large-scale detector that is capable of detecting up to $11$k classes while still outperforming YOLO-$9000$ and R-FCN-$3000$ by a large margin on coarse-grained classes.

Since there is no bounding box annotations for the $11$K classes in ImageNet, we demonstrate the qualitative results in Figure~\ref{fig:qualitative}. Overall, we find our model perform reasonably well. For example, our model is able to detect fine-grained animal species and human activities. More qualitative results and failure cases can be found in the supplemental file.

\vspace{-0.08in}
\section{Conclusion}

In this paper, we have proposed a semi-supervised based method to solve large-scale fine-grained object detection problem. Our method can achieve comparable results as state-of-the-art fully-supervised detector, by utilizing data from only a small number of fully-annotated coarse-grained classes and large scale weakly-annotated fine-grained classes. Our work not only establishes a new way of learning large scale detector, but also provides insights for large scale data collection and annotation.

There are a few future directions to explore. Currently we just use a simple two-level tree structure, and have not explored correlations within the coarse-grained or fine-grained set. Apparently, $11$k fine-grained classes should not be treated as flat and uncorrelated. we should consider utilized the hierarchical structures within the fine-grained classes. We could also try to explore systemic ways of find good fine-grained classes that can be reasonable well detected and also helpful for coarse-grained detection.



{\footnotesize
\bibliographystyle{ieee_fullname}
\bibliography{egbib}

\begin{thebibliography}{10}\itemsep=-1pt

\bibitem{Bahdanau2014}
Dzmitry Bahdanau, Kyunghyun Cho, and Yoshua Bengio.
\newblock Neural machine translation by jointly learning to align and
  translate.
\newblock {\em CoRR}, abs/1409.0473, 2014.

\bibitem{Bilen2016}
Hakan Bilen and Andrea Vedaldi.
\newblock Weakly supervised deep detection networks.
\newblock In {\em CVPR}, pages 2846--2854, 2016.

\bibitem{Bodla2017}
Navaneeth Bodla, Bharat Singh, Rama Chellappa, and Larry~S Davis.
\newblock Soft-nms - improving object detection with one line of code.
\newblock In {\em ICCV}, pages 5562--5570, 2017.

\bibitem{Chen2018}
Yanbei Chen, Xiatian Zhu, and Shaogang Gong.
\newblock Semi-supervised deep learning with memory.
\newblock In {\em ECCV}, 2018.

\bibitem{Dai2016}
Jifeng Dai, Yi Li, Kaiming He, and Jian Sun.
\newblock {R-FCN:} object detection via region-based fully convolutional
  networks.
\newblock In {\em NIPS}, pages 379--387, 2016.

\bibitem{Dai2017}
Jifeng Dai, Haozhi Qi, Yuwen Xiong, Yi Li, Guodong Zhang, Han Hu, and Yichen
  Wei.
\newblock Deformable convolutional networks.
\newblock In {\em ICCV}, pages 764--773, 2017.

\bibitem{Deng2009}
Jia Deng, Wei Dong, Richard Socher, Li{-}Jia Li, Kai Li, and Fei{-}Fei Li.
\newblock Imagenet: {A} large-scale hierarchical image database.
\newblock In {\em CVPR}, pages 248--255, 2009.

\bibitem{Everingham2010}
Mark Everingham, Luc J.~Van Gool, Christopher K.~I. Williams, John~M. Winn, and
  Andrew Zisserman.
\newblock The pascal visual object classes {(VOC)} challenge.
\newblock {\em International Journal of Computer Vision}, 88(2):303--338, 2010.

\bibitem{Gao2018}
Jiyang Gao, Jiang Wang, Shengyang Dai, Li{-}Jia Li, and Ram Nevatia.
\newblock {NOTE-RCNN:} noise tolerant ensemble {RCNN} for semi-supervised
  object detection.
\newblock {\em CoRR}, abs/1812.00124, 2018.

\bibitem{Ge2018}
Weifeng Ge, Sibei Yang, and Yizhou Yu.
\newblock Multi-evidence filtering and fusion for multi-label classification,
  object detection and semantic segmentation based on weakly supervised
  learning.
\newblock In {\em CVPR}, June 2018.

\bibitem{Girshick2015}
Ross~B. Girshick.
\newblock Fast {R-CNN}.
\newblock In {\em ICCV}, pages 1440--1448, 2015.

\bibitem{He2017}
Kaiming He, Georgia Gkioxari, Piotr Doll{\'{a}}r, and Ross~B. Girshick.
\newblock Mask {R-CNN}.
\newblock In {\em ICCV}, pages 2980--2988, 2017.

\bibitem{Kaiser2017}
Lukasz Kaiser, Ofir Nachum, Aurko Roy, and Samy Bengio.
\newblock Learning to remember rare events.
\newblock {\em CoRR}, abs/1703.03129, 2017.

\bibitem{openimages}
Ivan Krasin, Tom Duerig, Neil Alldrin, Vittorio Ferrari, Sami Abu-El-Haija,
  Alina Kuznetsova, Hassan Rom, Jasper Uijlings, Stefan Popov, Shahab Kamali,
  Matteo Malloci, Jordi Pont-Tuset, Andreas Veit, Serge Belongie, Victor Gomes,
  Abhinav Gupta, Chen Sun, Gal Chechik, David Cai, Zheyun Feng, Dhyanesh
  Narayanan, and Kevin Murphy.
\newblock Openimages: A public dataset for large-scale multi-label and
  multi-class image classification.
\newblock {\em Dataset available from
  https://storage.googleapis.com/openimages/web/index.html}, 2017.

\bibitem{Krishna2016}
Ranjay Krishna, Yuke Zhu, Oliver Groth, Justin Johnson, Kenji Hata, Joshua
  Kravitz, Stephanie Chen, Yannis Kalantidis, Li-Jia Li, David~A Shamma,
  Michael Bernstein, and Li Fei-Fei.
\newblock Visual genome: Connecting language and vision using crowdsourced
  dense image annotations.
\newblock 2016.

\bibitem{Lin2017}
Tsung{-}Yi Lin, Piotr Doll{\'{a}}r, Ross~B. Girshick, Kaiming He, Bharath
  Hariharan, and Serge~J. Belongie.
\newblock Feature pyramid networks for object detection.
\newblock In {\em CVPR}, pages 936--944, 2017.

\bibitem{Lin2014}
Tsung{-}Yi Lin, Michael Maire, Serge~J. Belongie, James Hays, Pietro Perona,
  Deva Ramanan, Piotr Doll{\'{a}}r, and C.~Lawrence Zitnick.
\newblock Microsoft {COCO:} common objects in context.
\newblock In {\em ECCV}, pages 740--755, 2014.

\bibitem{Liu2016}
Wei Liu, Dragomir Anguelov, Dumitru Erhan, Christian Szegedy, Scott~E. Reed,
  Cheng{-}Yang Fu, and Alexander~C. Berg.
\newblock {SSD:} single shot multibox detector.
\newblock In {\em ECCV}, pages 21--37, 2016.

\bibitem{Miller2016}
Alexander~H. Miller, Adam Fisch, Jesse Dodge, Amir{-}Hossein Karimi, Antoine
  Bordes, and Jason Weston.
\newblock Key-value memory networks for directly reading documents.
\newblock In {\em EMNLP}, pages 1400--1409, 2016.

\bibitem{Miller1995}
George~A. Miller.
\newblock Wordnet: {A} lexical database for english.
\newblock {\em Commun. {ACM}}, 38(11):39--41, 1995.

\bibitem{Papadopoulos2016}
Dim~P. Papadopoulos, Jasper R.~R. Uijlings, Frank Keller, and Vittorio Ferrari.
\newblock We don't need no bounding-boxes: Training object class detectors
  using only human verification.
\newblock In {\em CVPR}, pages 854--863, 2016.

\bibitem{Papadopoulos2017}
Dim~P. Papadopoulos, Jasper R.~R. Uijlings, Frank Keller, and Vittorio Ferrari.
\newblock Extreme clicking for efficient object annotation.
\newblock In {\em ICCV}, pages 4940--4949, 2017.

\bibitem{Redmon2016}
Joseph Redmon, Santosh~Kumar Divvala, Ross~B. Girshick, and Ali Farhadi.
\newblock You only look once: Unified, real-time object detection.
\newblock In {\em CVPR}, pages 779--788, 2016.

\bibitem{Redmon2017}
Joseph Redmon and Ali Farhadi.
\newblock {YOLO9000:} better, faster, stronger.
\newblock In {\em CVPR}, pages 6517--6525, 2017.

\bibitem{Redmon2018}
Joseph Redmon and Ali Farhadi.
\newblock Yolov3: An incremental improvement.
\newblock {\em CoRR}, abs/1804.02767, 2018.

\bibitem{Ren2017}
Shaoqing Ren, Kaiming He, Ross~B. Girshick, and Jian Sun.
\newblock Faster {R-CNN:} towards real-time object detection with region
  proposal networks.
\newblock {\em IEEE T-PAMI}, 39(6):1137--1149, 2017.

\bibitem{RFCN3K}
Bharat Singh, Hengduo Li, Abhishek Sharma, and Larry~S. Davis.
\newblock {R-FCN-3000} at 30fps: Decoupling detection and classification.
\newblock 2018.

\bibitem{Singh2018}
Bharat Singh, Mahyar Najibi, and Larry~S Davis.
\newblock {SNIPER}: Efficient multi-scale training.
\newblock In {\em NIPS}, 2018.

\bibitem{DOCK}
Krishna~Kumar Singh, Santosh~Kumar Divvala, Ali Farhadi, and Yong~Jae Lee.
\newblock {DOCK:} detecting objects by transferring common-sense knowledge.
\newblock In {\em ECCV}, pages 506--522, 2018.

\bibitem{Tang2017}
Peng Tang, Xinggang Wang, Xiang Bai, and Wenyu Liu.
\newblock Multiple instance detection network with online instance classifier
  refinement.
\newblock {\em CoRR}, abs/1704.00138, 2017.

\bibitem{Tang2018}
Peng Tang, Xinggang Wang, Angtian Wang, Yongluan Yan, Wenyu Liu, Junzhou Huang,
  and Alan~L. Yuille.
\newblock Weakly supervised region proposal network and object detection.
\newblock In {\em ECCV}, pages 370--386, 2018.

\bibitem{Tang2016}
Yuxing Tang, Josiah Wang, Boyang Gao, Emmanuel Dellandr{\'{e}}a, Robert~J.
  Gaizauskas, and Liming Chen.
\newblock Large scale semi-supervised object detection using visual and
  semantic knowledge transfer.
\newblock In {\em CVPR}, pages 2119--2128, 2016.

\bibitem{Tao2018}
Qingyi Tao, Hao Yang, and Jianfei Cai.
\newblock Exploiting web images for weakly supervised object detection.
\newblock {\em IEEE TMM}, 2018.

\bibitem{Tao2017}
Qingyi Tao, Hao Yang, and Jianfei Cai.
\newblock Zero-annotation object detection with web knowledge transfer.
\newblock In {\em ECCV}, 2018.

\bibitem{Uijlings2018}
Jasper R.~R. Uijlings, Stefan Popov, and Vittorio Ferrari.
\newblock Revisiting knowledge transfer for training object class detectors.
\newblock In {\em CVPR}, pages 1101--1110, 2018.

\bibitem{Uijlings2013}
Jasper R.~R. Uijlings, Koen E.~A. van~de Sande, Theo Gevers, and Arnold W.~M.
  Smeulders.
\newblock Selective search for object recognition.
\newblock {\em IJCV}, 104(2):154--171, 2013.

\bibitem{Wu2018}
Zhe Wu, Navaneeth Bodla, Bharat Singh, Mahyar Najibi, Rama Chellappa, and
  Larry~S. Davis.
\newblock Soft sampling for robust object detection.
\newblock {\em CoRR}, abs/1806.06986, 2018.

\bibitem{Yang2017}
Hao Yang, Joey~Tianyi Zhou, Jianfei Cai, and Yew{-}Soon Ong.
\newblock {MIML-FCN+:} multi-instance multi-label learning via fully
  convolutional networks with privileged information.
\newblock In {\em CVPR}, pages 5996--6004, 2017.

\bibitem{Zhang2018}
Yongqiang Zhang, Yancheng Bai, Mingli Ding, Yongqiang Li, and Bernard Ghanem.
\newblock W2f: A weakly-supervised to fully-supervised framework for object
  detection.
\newblock In {\em CVPR}, June 2018.

\bibitem{Zitnick2014}
C.~Lawrence Zitnick and Piotr Doll{\'{a}}r.
\newblock Edge boxes: Locating object proposals from edges.
\newblock In {\em ECCV}, pages 391--405, 2014.

\end{thebibliography}
}

\end{document}